\pdfoutput=1

\documentclass[11pt]{article}

\usepackage[]{EMNLP2023}

\usepackage{times}
\usepackage{latexsym}
\usepackage{graphicx}
\usepackage{amsmath}
\usepackage{nccmath}
\usepackage{amssymb}
\usepackage{mathtools}
\usepackage{multirow}
\usepackage{adjustbox}
\usepackage{todonotes}

\usepackage[T1]{fontenc}

\usepackage[utf8]{inputenc}

\usepackage{microtype}

\usepackage{inconsolata}

\DeclareUnicodeCharacter{2212}{-}

\title{Adapting Pre-trained Generative Models \\for Extractive Question Answering}

\author{Prabir Mallick \and Tapas Nayak \and Indrajit Bhattacharya  \\
        TCS Research, India \\
        \texttt{\{mallick.prabir,nayak.tapas,b.indrajit\}}@tcs.com\\}

\begin{document}
\maketitle
\begin{abstract}


Pre-trained Generative models such as BART, T5, etc. have gained prominence as a preferred method for text generation in various natural language processing tasks, including abstractive long-form question answering (QA) and summarization. However, the potential of generative models in extractive QA tasks, where discriminative models are commonly employed, remains largely unexplored. Discriminative models often encounter challenges associated with label sparsity, particularly when only a small portion of the context contains the answer. 
The challenge is more pronounced for multi-span answers.  
In this work, we introduce a novel approach that uses the power of pre-trained generative models to address extractive QA tasks by generating indexes corresponding to context tokens or sentences that form part of the answer. Through comprehensive evaluations on multiple extractive QA datasets, including MultiSpanQA, BioASQ, MASHQA, and WikiQA, we demonstrate the superior performance of our proposed approach compared to existing state-of-the-art models.

\end{abstract}
\section{Introduction}

An important subcategory of question-answering tasks is extractive question answering, where parts of a given context are selected as the answer to a question.
In many settings, this is considered more reliable than abstractive question answering \cite{Firsanova2021QuestionAS} which is more powerful in general but less explainable.  
The extractive question-answering task is primarily tackled using discriminative models. Specifically, for datasets featuring single-span factoid answers, such as SQuAD \cite{squad}, models such as \citet{zhang2021retrospective, yamada2020luke, zhang2020sg} identify the start and end positions of the answer span. Conversely, for datasets encompassing multi-span factoid answers, such as MultiSpanQA \cite{multispanqa_factoid} and BioASQ \cite{yoon2022sequence}, researchers have proposed discriminative models based on "BIO" tagging (`Begin', `Inside', `Outside'), 
which works for both single and multi-span answers.
In the case of long-form sentence-level QA datasets like MASHQA \cite{mashqa} and WikiQA \cite{wikiqa}, sentence classification models like MultiCo \cite{mashqa} have been employed. However, to date, the application of {\em generative} seq2seq models to address this {\em extractive} QA task remains unexplored.

The main challenge that we may hope to overcome using a generative approach is that of sparsity.
Our observations indicate that extractive question-answering tasks exhibit a high level of sparsity, where the answers comprise only a minuscule fraction of the tokens or sentences present in the given context (see Table \ref{tab:data_statistics}). For single-span answers, this sparsity does not pose a significant challenge, as models primarily focus on identifying the start and end positions of the answer span. Consequently, the loss function exclusively considers the answer-related context tokens, excluding the non-answer portion. However, in the case of multi-span answers utilizing "BIO" tagging, models encounter sparsity issues due to a large number of non-answer tokens being assigned "O" tags (Outside of the answer span). This sparsity challenge is also prevalent in sentence-level extractive QA datasets, such as MASHQA, where answer sentences are dispersed across multiple spans. State-of-the-art answer extraction models, such as MultiCo, employ sentence selection methods to identify the answer sentences. Given that answers can span multiple sentences across multiple spans, these discriminative sentence selection models similarly grapple with the sparsity of answers relative to the context.

\begin{table*}[t]
\small
\centering
\begin{tabular}{{p{0.95\textwidth}}}
\hline

{\bf Question}: What happens during a clinical trial for arthritis treatment? \\ \hline
{\bf Context}: \\
\textbf{1} \textcolor{red}{A clinical trial is a research study conducted with patients to evaluate a new arthritis treatment, drug, or device.} \\
\textbf{2} {The purpose of clinical trials is to find new and improved methods of treating arthritis.} \\
\textbf{3} {Clinical trials make it possible to apply the latest scientific and technological advances in arthritis to patient care.} \\
\textbf{4} \textcolor{red}{During a clinical trial, doctors use the best available arthritis treatment as a standard to evaluate new treatments.} \\
\textbf{5} \textcolor{red} {The new treatments are considered to be at least as effective or possibly more effective than the standard.} \\
\textbf{6} {New treatment options are first researched in a laboratory where they are carefully studied in the test tube and in animals.} \\
\textbf{7} \textcolor{red} {Only the treatments most likely to work are further evaluated in a small group of humans prior to applying them in a larger clinical trial. }\\
\textbf{8} {When a new arthritis treatment is studied for the first time in humans, it is not known exactly how it will work.} \\
\ldots \\
\textbf{17} {The researchers determine the best way to give the new treatment and how much of it can be given safely.} \\
\textbf{18} {Phase II clinical trials determine the effect of the research treatment on patients and usually the best dosage.} \\ \ldots
\\ \hline
{\bf Extracted Answer as Full Index (FI) Sequence}: 1 4 5 7 \\ 
{\bf Extracted Answer as Span Index (SI) Sequence}: (1 1) (4 5) (7 7) \\ \hline
{\bf Question}: When did India win the cricket world cup?
 \\ \hline
{\bf Context}: \\
\textbf{0} The \textbf{1} Indian \textbf{2} cricket \textbf{3} team \textbf{4} are \textbf{5} two \textbf{6} times \textbf{7} World \textbf{8} Champions \textbf{9} . \textbf{10} In \textbf{11} addition \textbf{12} to \textbf{13} winning \textbf{14} the \textbf{15} \textcolor{red}{1983} \textbf{16} Cricket \textbf{17} World \textbf{18} Cup \textbf{19}, \textbf{20} they \textbf{21} triumphed \textbf{22} over \textbf{23} Sri \textbf{24} Lanka \textbf{25} in \textbf{26} the \textbf{27} \textcolor{red}{2011} \textbf{28} Cricket \textbf{29} World \textbf{30} Cup \textbf{31} on \textbf{32} home \textbf{33} soil \textbf{34} . \textbf{35} They \textbf{36} were \textbf{37} also \textbf{38} runners \textbf{39} - \textbf{40} up \textbf{41} at \textbf{42} the \textbf{43} 2003 \textbf{44} Cricket \textbf{45} World \textbf{46} Cup \textbf{47}, \textbf{48} and \textbf{49} semifinalists \textbf{50} thrice \textbf{51} (\textbf{52} 1987 \textbf{53}, \textbf{54} 1996 \textbf{55} and \textbf{56} 2015 \textbf{57}) \textbf{58} . \ldots \ldots \ldots
\textbf{94} India \textbf{95}'s \textbf{96} historical \textbf{97} win \textbf{98} - \textbf{99} loss \textbf{100} record \textbf{101} at \textbf{102} the \textbf{103} cricket \textbf{104} world \textbf{105} cup \textbf{106} is \textbf{107} 46 \textbf{108} - \textbf{109} 27 \textbf{110}, \textbf{111} with \textbf{112} 1 \textbf{113} match \textbf{114} being \textbf{115} tied \textbf{116} and \textbf{117} another \textbf{118} one \textbf{119} being \textbf{120} abandoned \textbf{121} due \textbf{122} to \textbf{123} rain \textbf{124}. \ldots
\\ \hline
{\bf Extracted Answer as Full Index (FI) Sequence}: 15 27 \\ 
{\bf Extracted Answer as Span Index (SI) Sequence}: (15 15) (27 27) \\ \hline
\end{tabular}
\caption{Illustration of the task of extracting reference answers using two examples. The first example is from MASHQA depicting sentence-level tasks and the second example is from MultiSpanQA depicting token-level tasks. These two examples show the representation of the context, the answer spans, and two different representations of the answer spans using indexes. For the span index (SI) sequence, each pair denotes the beginning and end indexes of the span. Indexes in the context are shown in bold, answer spans in red, and parentheses are added for span index sequence for ease of illustration.}
\label{tab:example}
\end{table*}

The sparsity challenge encountered in extractive question answering is less daunting for generative approaches, as they explicitly model what is likely (via likelihood) rather than what is unlikely. Moreover, the remarkable performance exhibited by large pre-trained generative seq2seq models such as BART, T5, etc. in various tasks has been well-documented in recent years \cite{rebel, retrieve_gen}. However, the application of generative seq2seq models to an extractive task raises two key questions: What should the model generate, and does this unnecessarily complicate the task? To address these concerns, we propose a novel approach: generating the indexes of context tokens or sentences that form part of the extractive answer. By adopting this generative strategy, we effectively restrict the output space, facilitating the learning of a distribution over a reduced set of possibilities. Additionally, the burden of training is alleviated through fine-tuning large pre-trained models. Notably, to the best of our knowledge, no prior work has employed index generation via generative models for extractive question answering. We demonstrate the superiority of our generative approach over state-of-the-art answer extraction models. A key advantage of our proposed approach lies in its simplicity and applicability to any multi-span extractive task. Through comprehensive evaluation on five extractive QA datasets, we establish its superiority over existing customized models designed for specific datasets\footnote{Any resources related to this work will be made available at \url{https://github.com/prabirmallick/GenAI4EQA}}. 

\section{Adaptation of Generative Model for Extractive Question Answering} 
\label{sec:model}
We now formalize the sentence-level answer extraction task and propose a novel approach for it. This can easily be extended for the token-level answer extraction tasks as well. We are given a context $c$ and a question $q$.
The context is a sequence of sentences $\{s_1,s_2,\ldots,s_{n}\}$, where $n$ is the number of sentences in $c$.
Each sentence $s_i$ and similarly the question $q$ is a sequence of tokens.
Each sentence is associated with a binary variable $a_i$ to indicate whether it is part of the extractive answer for $q$.
The answer sentences, with $a_i=1$ may form one or more spans in the context.

\paragraph{Generative Seq2Seq Model for Answer Extraction:}
A generative sequence-to-sequence model, such as BART~\cite{bart} and T5 \cite{T5}, uses chain rule and models the probability of each token $o_i$ in the output sequence $o$, conditioned on the input sequence $x$ and the previously generated output tokens $o_{<i}$: $\prod_{i=i}^n P(o_i \mid o_{<i},x)$. 
The model is trained by maximizing the log-likelihood of the output tokens in the training data.

Our goal is to identify the answer sentences in the input context using a generative model. 
An {\em indirect} approach is to first generate an answer and then use it to identify spans from the context \cite{xu2021attention}.
We investigate more {\em direct} approaches for `generating the extractive answer'.
The simplest direct approach is to generate the answer token by token by learning to copy sentences from the input to the output. 
But this requires extremely large volumes of data to learn. 
We investigate a more compressed representation of the extractive answer whose generation can be learned more efficiently.

We propose to generate {\em the indexes of the answer sentences} in the context. 
We explore two different strategies to generate the answer sentence indexes:- 

(II) \textbf{Full Index (FI)} Sequence Generation: In this approach, the output sequence is the sequence of the indexes of all the sentences that are in the answer, i.e., $a_i=1$.

(II) \textbf{Span Index (SI)} Sequence Generation: A span of answer text in a context can be more compactly represented with the indexes of the first and last elements of the answer span. As a span-based representation of answers, we use the indexes of the first and last sentences of the answer span. For multi-span answers, we represent the sequence of spans, each using their corresponding start and end sentence indexes.


To facilitate this index-based generation, we modify the input context $c$ by inserting the sentence index number before each sentence in the context. We include an example in Table \ref{tab:example} to illustrate our approach. As generation of the indexes are not constrained in generative models, we appropriately post-process the output to obtain valid answer sequences (see subsection \ref{index_post_process}). To extend this model for token-level tasks, we just replace the sentence indexes with token indexes in the context and in the output. We use BART-base (BART$_b$) and BART-large (BART$_l$) \cite{bart} as representative of pre-trained generative models for our experiments. 


\subsection{Inference-time Index Post-processing}
\label{index_post_process}

The use of an index-based representation for the answer has the advantage of constraining the output space, resulting in significantly shorter sequences. However, it's essential to note that this approach doesn't inherently guarantee that the output will constitute a valid extractive answer. During the inference phase, indexes may be generated in a non-sequential order, duplicates may appear, and, in the worst-case scenario, out-of-range indexes can emerge. To address these issues in the context of full index generation (FI), we implement a post-processing step. This step involves sorting the generated indexes and removing any that fall outside the valid range.

The challenge becomes more pronounced when dealing with span index (SI) generation. In this case, the potential for invalid sequences multiplies, including scenarios where the sequence length is odd, the start index of a span exceeds the end index, spans intersect or encompass each other, or spans extend beyond the valid range. To address these complexities, our post-processing strategy involves: (i) Pruning unpaired last indexes. (ii) Removing spans that are invalid or out of range. (iii) Merging overlapping spans. It's noteworthy that, in practice, the occurrence of invalid indices is relatively rare, accounting for less than 1\% of generated indices. We carefully handle such invalid indices during post-processing, retaining only the valid ones to obtain the final answer.

\section{Experiments}

\subsection{Datasets}


As our proposed generative approach produces a sequence in the output, we choose datasets that have multiple spans as answers. For factoid answer extraction, we use \textbf{MultiSpanQA} ~\cite{multispanqa_factoid} and \textbf{BioASQ} for experiments. \textbf{MultiSpanQA} contains only multi-span answers and does not include any single-span answers. The answer labels for the test set of this dataset are not publicly available. We need to submit the predictions on the test to the leaderboard team to obtain the test performance on MultiSpanQA. 
{\bf BioASQ} ~\cite{yoon2022sequence} \textbf{BioASQ7b}, and \textbf{BioASQ8b} is a benchmark for biomedical question answering with list-type questions with multiple extractive factoid answers. 

For long-form QA, we use \textbf{MASHQA} \cite{mashqa} dataset from the medical domain. Each answer in this dataset consists of one or more sentences from the context but these answer sentences may not be continuous in the context. \textbf{WikiQA} \cite{wikiqa} is another sentence-level extractive QA dataset but here questions have just a single sentence answer. Detailed statistics of the various datasets used in our experiments are recorded in Table \ref{tab:data_statistics}. 

\begin{table*}[ht]
\small
\centering
\begin{tabular}{l|c|c|c|c|c|c|c}
\hline
Dataset     & Answer Type    & Multispan ? & Train        & Validation & Test      & \begin{tabular}[c]{@{}c@{}}Label \\ Sparsity (\%)\end{tabular} & \begin{tabular}[c]{@{}c@{}}\% Context \\ Trimmed\end{tabular} \\ \hline
MASHQA      & Sentence-level & Yes         & 19,895/4,250 & 2,669/474  & 2,582/473 & 17-18                                                               &  67                                                             \\ 
WikiQA      & Sentence-level & No          & 565/0        & 64/0       & 146/0     & 2-3                                                               &  10                                                             \\ 
MultiSpanQA & Token-level    & Yes         & 0/5,230      & 0/653      & NA/NA     & 3-4                                                               &  1                                                             \\ 
BioASQ7b    & Token-level    & Yes         & 3610/3610    & 393/393    & 393/393   & 2-3                                                               &  0                                                             \\ 
BioASQ8b    & Token-level    & Yes         & 3914/3914    & 383/383    & 383/383   & 2-3                                                               &  0                                                             \\ \hline
\end{tabular}
\caption{Statistics of MASHQA, WikiQA, MultiSpanQA, and BioASQ datasets. n/m denotes single-span/multi-span answer counts. In the MultiSpanQA dataset, the gold labels of the test dataset are not available (NA). We need to submit our predicted answers to the MultiSpanQA leaderboard to obtain the scores on their test dataset.}
\label{tab:data_statistics}
\end{table*}

\textbf{Context Trimming:} We utilize the BART model as our generative framework, which comes with a maximum token capacity of 1,024. In some cases, to accommodate the context appropriately, we must truncate a portion of it. To ensure that the resulting input still encompasses the entire answer, we retain a maximum of 1,024 tokens from the original context. To achieve this, we extract the complete answer span from the original context and extend it both to the left and right, crafting a contiguous sequence of 1,024 tokens. Any instances where the answer span exceeds this 1,024-token limit are omitted. This particular situation arises for a relatively small fraction (10\%) of multi-span answers, where the answer sentences are dispersed widely within an extensive context. In Table \ref{tab:data_statistics}, we provide information on the percentage of sentences removed during this trimming process for various datasets. Notably, the MASHQA dataset is notably affected, with approximately 67\% of its sentences needing removal to fit within the confines of the BART encoder.

\textbf{Label Sparsity:} In Table \ref{tab:data_statistics}, we incorporate a measure of label sparsity for the QA datasets following the context trimming process. This measure reveals the percentage of sentences or tokens within the context that are relevant to the answer. Notably, in the MASHQA dataset, approximately 17-18\% of the context sentences are part of the answer, whereas in other datasets it is around 2-4\% a significantly lower figure compared to MASHQA. With this kind of imbalance between the answer part and the non-answer part of the context, every sentence or token must be classified by the discriminative models. Consequently, this label imbalance poses a challenge for the discriminative models, as they grapple with the need to assign labels to a wide array of context elements in a nuanced manner.

\subsection{Evaluation Metrics}
\label{app:evaluation_metric}
We use sentence-level precision, recall, and F1 scores for the sentence-level QA datasets MASHQA and WikiQA. Similarly, we use token-level precision, recall, and F1 score for the BioASQ dataset. But for the MultiSpanQA dataset, we report precision, recall, and F1 scores based on exact match (EM) and partial match (PM). In the EM-based F1 score, all the spans of the ground truth answer must match with the predicted answer spans.

\subsection{Baseline Models}

\textbf{(i)} We use multiple pre-trained language models such as BERT \cite{devlin2018bert}, RoBERTa, BioBERT \cite{lee2020biobert}, PubMedBERT \cite{pubmedbert}, XLNet \cite{xlnet} as baselines. For multi-span factoid answers, we use a `BIO' tagging head on top of these models, and for sentence-level extraction, we use a sentence classifier head. 

\textbf{(ii)} We fine-tune a BART-base \cite{bart} seq2seq model that directly generates the token sequence in the answer, which we call BART\_Text or BART\_T in short. We link back the generated answers to context sentences for evaluation under the extractive paradigm (see details in \ref{app:link_back}). 

\textbf{(iii)} MultiCo \cite{mashqa} is another sentence-level classification model that encodes a question and context pair using XLNet~\cite{xlnet} and classifies each context sentence as part of the answer or not. It uses sparsified inter-sentence attention for each sentence to get weights over other context sentences. 

\textbf{(iv)} As a few-shot baseline, we employed the \textbf{Flan-T5} large model \cite{Chung2022ScalingIL} with eight examples. However, while attempting to generate indexes using this model, we found it to be unsuccessful. Consequently, we directly generated the answer in the few-shot setting for factoid answers. For sentence-level answers, we mapped the generated answer back to the corresponding context sentences (see details in \ref{app:link_back}).

\textbf{(v)} \textbf{LIQUID} \cite{Lee2023LIQUIDAF} is an answer generation framework that utilizes unlabelled corpora to generate high-quality synthetic datasets for various QA tasks. By fine-tuning RoBERTa-base or RoBERTa-large \cite{roberta} with a `BIO' tagging head on both the synthetic dataset and task-specific dataset, LIQUID achieves state-of-the-art performance on the MultiSpanQA and BioASQ datasets.

\subsection{Linking back Abstractive Answer to Context Sentences}
\label{app:link_back}

We employ a token overlap mechanism to align the abstractive long-form content generated by models such as BART/Flan-T5 with the corresponding context sentences. It's worth noting that extractive answers can encompass varying numbers of spans. To perform this alignment, we leverage spaCy\footnote{https://spacy.io/} to calculate the token-wise overlap between each context sentence and the generated answer. Subsequently, we pinpoint the context sentences that exhibit a substantial token overlap with the generated answer. It's important to highlight that the quantity of context sentences may differ for each answer. We select the sentence with the highest token overlap and, in addition, include those sentences with token overlap values close to that of the most similar sentence. This approach draws parallels with the concept of identifying a knee point in a dataset, akin to the knee detection problem. The deviation in token overlap from the most similar sentence is employed as a hyper-parameter for fine-tuning the link-back algorithm.

\subsection{Parameter Settings}
\label{app:param_settings}

We use pre-trained BART-base (BART$_b$) and BART-large (BART$_l$) as our generative model. 
We train our models with a batch size of 8 and update the model parameters using AdamW optimizer ~\cite{adamw} with learning rate $2\times10^{-5}$ and weight decay $1\times10^{-4}$. 
We use early stopping if there is no improvement on the validation set for the last 5 evaluations.
All our experiments are performed on an NVIDIA MiG A100 with 60 GB RAM and 20 GB GPU memory. 
We report an average of three runs for our proposed framework. 

BART restricts maximum encoder and decoder lengths to 1024 tokens. 
The contexts are often longer than this encoder limit, particularly for the MASHQA dataset. 
To fit the context in BART, we trim these contexts, while ensuring that the trimmed context includes the entire gold-standard extracted answers. 
All evaluations for all models including baselines are reported on the trimmed datasets.

\section{Experimental Results}

In our initial experiments, we focus on QA datasets containing short answer spans, such as MultiSpanQA and BioASQ, and we present the corresponding performance in Tables \ref{tab:multispanqa} and \ref{tab:bioasq}. Notably, we observed that both our proposed full index sequence generation and span index sequence generation methods yield comparable results on these datasets. Specifically, our BART\_FI$_l$ model outperforms the LIQUID model \cite{Lee2023LIQUIDAF} by 1\% in terms of F1 score based on partial match evaluation on MultiSpanQA. Moreover, on the BioASQ8b dataset, both our BART\_SI$_l$ and BART\_FI$_l$ models achieve new state-of-the-art (SOTA) performance, surpassing the previous SOTA LIQUID$_l$ model by an impressive margin of 4\%. Additionally, our model achieves performance on the BioASQ7b dataset that is very close to the SOTA performance of LIQUID$_l$.

\begin{table*}[ht]
\centering
\begin{tabular}{l|ccc|ccc}
\hline
 & \multicolumn{3}{c|}{Exact Match} & \multicolumn{3}{c}{Partial Match}  \\ \hline
 Model & Prec. & Rec. & F1 & Prec. & Rec. & F1 \\ \hline
 FLAN-T5$_{large}$ & 0.45 & 0.23 & 0.30 & 0.72 & 0.53 & 0.61  \\ 
BERT$_{base}$ & 0.58 & 0.61 & 0.59 & 0.80 & 0.73 & 0.76 \\
BART\_Text$_{base}$ & 0.59 & 0.61 & 0.60 & 0.80 & 0.77 & 0.78 \\ 
LIQUID-RoBERTa$_{base}$ & 0.66 & 0.69 & 0.67 & 0.81 & 0.81 & 0.81  \\ 
LIQUID-RoBERTa$_{large}$ & 0.75 & 0.68 & \textbf{0.71} & 0.85 & 0.77 & 0.81 \\ \hline 
BART\_SI$_{base}$ & 0.62 & 0.61 & 0.61 & 0.79 & 0.75 & 0.77  \\
BART\_FI$_{base}$ & 0.61 & 0.62 & 0.61 & 0.78 & 0.76 & 0.77 \\
BART\_SI$_{large}$ & 0.67 & 0.69 & 0.68 & 0.81 & 0.82 & 0.81  \\
BART\_FI$_{large}$ & 0.66 & 0.70 & 0.68 & 0.80 & 0.85 & \textbf{0.82}  \\ \hline

\end{tabular}%
\caption{Performance comparison of our proposed model against the SOTA baselines on MultiSpanQA.}
\label{tab:multispanqa}
\end{table*}

\begin{table*}[ht]
\centering
\begin{tabular}{l|lll|lll}
\hline
                            & \multicolumn{3}{c|}{BioASQ7b} & \multicolumn{3}{c}{BioASQ8b} \\ \hline
Model                       & Prec.        & Rec.        & F1       & Prec.        & Rec.       & F1       \\ \hline
FLAN-T5$_{large}$                & 0.23     & 0.45     & 0.31    & 0.16     & 0.40    & 0.23    \\
BART\_Text$_{base}$  & 0.25     & 0.41     & 0.31    & 0.22     & 0.41    & 0.29    \\ 
BioBERT$_{base}$ & 0.42     & 0.58     & 0.45    & 0.39     & 0.59    & 0.44    \\
PMBERT$_{base}$ & 0.43     & 0.63     & 0.48    & 0.38     & 0.59    & 0.43    \\
LIQUID-RoBERTa$_{base}$ & 0.41     & 0.61     & 0.49    & 0.37     & 0.56    & 0.44    \\
LIQUID-RoBERTa$_{large}$  & 0.45     & 0.64     & \textbf{0.53}    & 0.39     & 0.59    & 0.47    \\ \hline
BART\_SI$_{base}$              & 0.44     & 0.56     & 0.49    & 0.42     & 0.50    & 0.46    \\ 
BART\_FI$_{base}$               & 0.43     & 0.58     & 0.49    & 0.42     & 0.51    & 0.46    \\
BART\_SI$_{large}$                & 0.46     & 0.59     & 0.52    & 0.46     & 0.56    & \textbf{0.51}    \\ 
BART\_FI$_{large}$                & 0.46     & 0.59     & 0.52    & 0.46     & 0.55    & \textbf{0.51}    \\
\hline
\end{tabular}%
\caption{Performance comparison of our proposed method against the SOTA baselines on BioASQ 7b and 8b datasets. PMBERT refers to PubMedBERT.}
\label{tab:bioasq}
\end{table*}

Subsequently, we conduct experiments on sentence-level long-form QA datasets, namely MASHQA and WikiQA, and present the outcomes in Table \ref{tab:mashqa_wikiqa}. Remarkably, our BART-large models, namely BART\_SI$_l$ and BART\_FI$_l$, achieve a noteworthy improvement in performance compared to previous state-of-the-art (SOTA) models. Specifically, on the MASHQA dataset, both BART\_SI$_l$ and BART\_FI$_l$ models attained a 3-4\% higher F1 score compared to the previous SOTA XLNet model. Similarly, on the WikiQA dataset, our BART-large models outperformed the previous SOTA Flan-T5 model by 3\% in terms of F1 score. These results unequivocally demonstrate that our proposed adaptation of the pre-trained generative model surpasses the performance of baseline models in the sentence-level answer extraction task, without necessitating any task-specific modifications to the model architecture.

\begin{table*}[ht]
\centering
\begin{tabular}{l|lll|lll}
\hline
Model        & \multicolumn{3}{c|}{MASHQA} & \multicolumn{3}{c}{WikiQA} \\ \hline
             & Prec.        & Rec.       & F1      & Prec.       & Rec.       & F1     \\ \hline
BART\_{Text}$_{base}$       & 0.59     & 0.32    & 0.41    & 0.47    & 0.35    & 0.40    \\ 
XLNet$_{base}$        & 0.61     & 0.74    & 0.67    & 0.49    & 0.51    & 0.50    \\
BERT$_{base}$         & 0.42     & 0.63    & 0.50     & 0.48    & 0.56    & 0.52   \\
RoBERTa$_{base}$      & 0.48     & 0.62    & 0.54    & 0.56    & 0.54    & 0.55   \\

MultiCo-XLNet$_{base}$      & 0.61     & 0.73    & 0.66    & 0.57    & 0.57    & 0.57   \\

FLAN-T5$_{large}$  & 0.62     & 0.22    & 0.33    & 0.68     & 0.68     & 0.68    \\ \hline
BART\_SI$_{base}$ & 0.67     & 0.69    & 0.68    & 0.63    & 0.63     & 0.63    \\ 
BART\_FI$_{base}$  & 0.65     & 0.71    & 0.68    & 0.64    & 0.64     & 0.64   \\
BART\_SI$_{large}$ & 0.65     & 0.75    & 0.70    & 0.70     & 0.70    & 0.70    \\ 
BART\_FI$_{large}$  & 0.66     & 0.77    & \textbf{0.71}    & 0.71     & 0.71     & \textbf{0.71}    \\ \hline

\end{tabular}%
\caption{Performance comparison of our proposed model against SOTA baselines on MASH-QA and WikiQA in terms of sentence level Precision, Recall and F1 scores.}
\label{tab:mashqa_wikiqa}
\end{table*}

We include the previous SOTA performance and our best F1 score across the five datasets in Table \ref{tab:sota_comparison}. We see that our proposed model achieved new SOTA on four of these five datasets and performed competitively on the remaining one dataset. In summary, the experimental findings presented above provide compelling evidence that the index sequence generation approach consistently outperforms specialized state-of-the-art models across a wide range of extractive QA tasks and datasets, without the need for task-specific customization. It is worth noting that previous state-of-the-art models do not consistently deliver optimal performance across all five datasets. In contrast, our proposed model demonstrates consistent performance across all these datasets, showcasing its remarkable generalization capability.

\begin{table*}[ht]
\centering
\begin{tabular}{l|c|c|c|c|c}
\hline
              & MultiSpanQA & BioASQ7b & BioASQ8b & MASHQA & WikiQA \\ \hline
Previous SOTA & 0.81    & \textbf{0.53}    &  0.47  & 0.67  & 0.68       \\ 
Our Best      & \textbf{0.82}    &  0.52   &  \textbf{0.51}  & \textbf{0.71}  & \textbf{0.71}       \\ \hline
\end{tabular}
\caption{F1 score comparison of best performance achieved by our proposed framework against the previous SOTA across five datasets.}
\label{tab:sota_comparison}
\end{table*}


\begin{table*}[ht]
\centering
\begin{tabular}{l|ccc|ccc}
\hline
            & \multicolumn{3}{c|}{BART\_FI$_{base}$}                        & \multicolumn{3}{c}{BART\_FI$_{large}$}                        \\ \hline
 & \multicolumn{1}{c|}{With Index} & \multicolumn{1}{c|}{Without Index} & $\uparrow (\%)$ & \multicolumn{1}{c|}{With Index} & \multicolumn{1}{c|}{Without Index} &  $\uparrow (\%)$ \\ \hline
MASHQA      & \multicolumn{1}{c|}{0.68} & \multicolumn{1}{c|}{0.63} & 5\% & \multicolumn{1}{c|}{0.71} & \multicolumn{1}{c|}{0.64} & 7\% \\ 
WikiQA      & \multicolumn{1}{c|}{0.64} & \multicolumn{1}{c|}{0.37} & 27\% & \multicolumn{1}{c|}{0.71} & \multicolumn{1}{c|}{0.36} & 35\% \\ 
MultiSpanQA & \multicolumn{1}{c|}{0.77} & \multicolumn{1}{c|}{0.52} & 25\% & \multicolumn{1}{c|}{0.82} & \multicolumn{1}{c|}{0.53} & 29\% \\ 
BioASQ7b    & \multicolumn{1}{c|}{0.49} & \multicolumn{1}{c|}{0.05} & 44\% & \multicolumn{1}{c|}{0.52} & \multicolumn{1}{c|}{0.06} & 46\% \\ 
BioASQ8b    & \multicolumn{1}{c|}{0.46} & \multicolumn{1}{c|}{0.06} & 40\% & \multicolumn{1}{c|}{0.51} & \multicolumn{1}{c|}{0.07} & 44\% \\ \hline
\end{tabular}
\caption{Ablation of our proposed model when trained with or without the sentence or token index in the input context. For MultiSpanQA, we report the partial match F1 score here. $\uparrow (\%)$ refers to the increase in F1 score in absolute percentage when indexes are added in the context.}
\label{tab:ablation}
\end{table*}

\subsection{Ablation Study}

Table \ref{tab:ablation} presents the ablation study of our model. Since we have limited flexibility in modifying the BART model itself, the only ablation we considered is removing the index tokens from the context and generating the answer indexes accordingly. From the results in Table \ref{tab:ablation}, we observe that the performance of both BART-base and BART-large models is relatively consistent on each dataset when indexes are not included in the context. When indexes are not included in the context, the BART-large model does not give any significant performance boost over the BART-base model on any of the datasets. This suggests that these models struggle to comprehend the meaning of the output sequence in the absence of index tokens in the context. From Table \ref{tab:ablation}, we can clearly observe that incorporating the index numbers into the context significantly enhances the performance of BART-base and BART-large models. 


\section{Related Work}

\paragraph{QA Tasks and Datasets:}
The early QA tasks involved open-domain reading-comprehension-style questions with factoid answers spanning a few words in the context.
The contexts were typically from general Wikipedia articles, news sites, and other web pages.
SQuAD~\cite{squad}, MS-MARCO~\cite{msmarco}, NewsQA~\cite{newsqa}, TriviaQA~\cite{triviaqa}, and SearchQA~\cite{searchqa} are some of the popular datasets.
Similar QA datasets also exist for the medical and science domains.  
emrQA ~\cite{pampari2018emrqa} has questions from healthcare clinical notes, while OpenBookQA \cite{sciqa_1} and ARC \cite{sciqa_2} have questions from elementary science texts.


More recent datasets such as Natural Questions~\cite{google_nq} and ELI5~\cite{eli5} contain questions with long answers, typically 2-5 sentences. 
In ELI5, which is based on community question-answering forums, the answers are not extractive but abstractive.
MASHQA~\cite{mashqa} has extractive, long-form, multi-span answers to questions about health and medicine. 


Few datasets have multi-span extractive answers.
Some are for multiple factoid answers spread across multiple sentences in the context~\cite{multispanqa_factoid,yoon2022sequence,xu2021evaluation}.
Answers in MASHQA~\cite{mashqa} are long-form and multi-span.



\paragraph{QA Approaches:}
With the arrival of large QA datasets such as SQuAD, deep pointer-network-based span extraction models came to the forefront of question-answering~\cite{chen2017reading, seo2016bidirectional, xiong2016dynamic}. 
Fine-tuning pre-trained language models such as BERT \cite{devlin2018bert}, XLNet \cite{xlnet} and RoBERTa \cite{roberta} for span extraction is the state-of-the-art for factoid question answering. 
But this approach does not work well for long-form and multi-span answers.
For long-form answers that span one or more sentences (continuous or non-continuous), sentence selection models have been shown to perform better~\cite{mashqa} than the span extraction models.

\paragraph{Generative Models for Extractive Tasks:}
Large generative pre-trained language models (PLMs) such as BART \cite{bart}, T5 \cite{T5} are mostly used for text generation tasks such as abstractive QA, abstractive summarization, etc. But in recent times, they are explored for many extractive tasks as well such as relation extraction \cite{rebel}, passage retrieval \cite{retrieve_gen}, etc.
\citet{xu2021attention} propose an approach to {\em indirectly} obtain an extractive span for factoid-style answers from a generated answer by recovering context sentences using decoder cross-attention patterns.
For long contexts with sentences, \citet{chowdhury2021everything} use a generative strategy for the sentence reordering task.
Generating sentence indexes has also been used as a component in a larger architecture for multi-hop QA~\cite{mhop}.
However, to the best of our knowledge, there is no work on {\em directly} using generative seq2seq models for long-form multi-span answer extraction.

\section{Conclusion}

In this work, we introduce a novel approach for extractive question-answering by leveraging a pre-trained generative language model and fine-tuning it to generate indexes of answer tokens or sentences. Discriminative models often necessitate dataset-specific customizations to achieve satisfactory performance due to the varying nature of label sparsity in such tasks. In contrast, we demonstrate that generative models can be readily adapted to address this challenge by generating indexes of tokens instead of directly generating the tokens themselves. Through empirical evaluations, our proposed model surpasses specialized state-of-the-art baselines across a range of diverse extractive QA benchmark datasets, showcasing its superior performance and effectiveness.

\section{Limitations}

Although we have highlighted the novelty, significance, and strengths of our proposed approach, it is essential to acknowledge some limitations. One limitation stems from the length restrictions imposed by generative encoders and decoders, such as BART, on input and output sequences. This constraint poses challenges when accommodating very long contexts, despite our efforts to address this issue within the scope of this paper. Future research should focus on developing improved solutions to handle longer contexts effectively.

Furthermore, a limitation of employing token-level indexing for multi-span factoid questions is the substantial amount of additional information required to represent the context. The need to insert an index for each token in the context can be disadvantageous, particularly in scenarios where there are limitations on the context length for large language models. This drawback highlights the importance of exploring alternative representations or encoding mechanisms that can effectively capture multi-span factoid questions while minimizing the impact on context length limitations.

Also, as we are posing an extractive task in a generative style, this may introduce the problem of exposure bias. Since exposure bias is a general problem for any auto-regressive model, any general solution to this problem is applicable to our proposed framework as well.


\section{Ethics Statement}

Our work does not have any ethical issues or obvious risks.

\bibliography{anthology,custom}
\bibliographystyle{acl_natbib}

\end{document}